\pdfoutput=1
\documentclass{article}

 \usepackage[main, final]{neurips_2026}
\usepackage[utf8]{inputenc} %
\usepackage[T1]{fontenc}    %
\usepackage[hidelinks]{hyperref}       %
\usepackage{url}            %
\usepackage{booktabs}       %
\usepackage{amsfonts}       %
\usepackage{nicefrac}       %
\usepackage{microtype}      %
\usepackage{xcolor}         %
\usepackage{graphicx}   %
\usepackage{booktabs}   %
\usepackage{multirow}   %
\usepackage{tcolorbox}
\usepackage{titlesec}
\usepackage{titletoc}
\usepackage{lineno}
\usepackage[table]{xcolor}
\usepackage{amsmath}
\usepackage{xspace}
\usepackage{wrapfig}
\usepackage{lipsum}
\usepackage[ruled,vlined]{algorithm2e}
\usepackage{float}
\usepackage{wrapfig}
\usepackage{pgfplots}
\pgfplotsset{compat=1.18}

\newcommand{\method}{dMoE\xspace}

\definecolor{map_red}{RGB}{239,99,75}
\definecolor{map_blue}{RGB}{99,113,250}
\definecolor{map_green}{RGB}{0,180,139}
\definecolor{map_yellow}{RGB}{229,157,35}
\definecolor{map_gray}{RGB}{165,165,165}
\definecolor{link}{RGB}{229,158,221}
\definecolor{mygreen}{RGB}{93,173,85}
\definecolor{myred}{RGB}{192,57,43}

\hypersetup{
  colorlinks=true,
  linkcolor={red!70!black},
  citecolor=link,
  urlcolor=cyan,
}

\title{\method: dLLMs with Learnable Block Experts}

\author{%
  Sicheng Feng, Zigeng Chen, Gongfan Fang, Xinyin Ma, Xinchao Wang\thanks{Corresponding Author} \\
  National University of Singapore \\
  \texttt{fengsicheng@u.nus.edu}, \texttt{xinchao@nus.edu.sg} \\
}

\begin{document}

\maketitle

\begin{abstract}
Diffusion Large Language Models (dLLMs) have recently emerged as a promising alternative to autoregressive models, offering competitive performance while naturally supporting parallel decoding. 
However, as dLLMs are increasingly integrated with Mixture-of-Experts (MoE) architectures to scale model capacity, a fundamental mismatch arises between block parallel decoding and token-level expert selection. Specifically, each dLLM forward pass processes multiple tokens with bidirectional dependencies, whereas conventional MoE layers route each token independently. This mismatch substantially increases the number of uniquely activated experts, making inference increasingly memory-bound.
To address this, we propose \method, a simple yet effective block-level MoE framework. The central idea of \method is to aggregate token-level expert distributions within each block into a unified block-level expert distribution, which is then used to guide block expert routing in a coherent manner. In this way, \method substantially reduces the number of uniquely activated experts during inference without sacrificing performance, thereby mitigating the memory-bound bottleneck.
Extensive experiments across a variety of benchmarks demonstrate the effectiveness of \method. On average, \method reduces the number of uniquely activated experts from 69.5 to 16.6 while retaining 99.55\% of the original performance. Meanwhile, it reduces memory usage by 74.36\% to 76.78\% and achieves $1.36\times$ to $2.04\times$ end-to-end latency speedup.
Code is available at: \url{https://github.com/fscdc/dMoE}.
\end{abstract}

\begin{figure*}[ht]
\centering
\vspace{-3mm}
  \includegraphics[width=\linewidth]{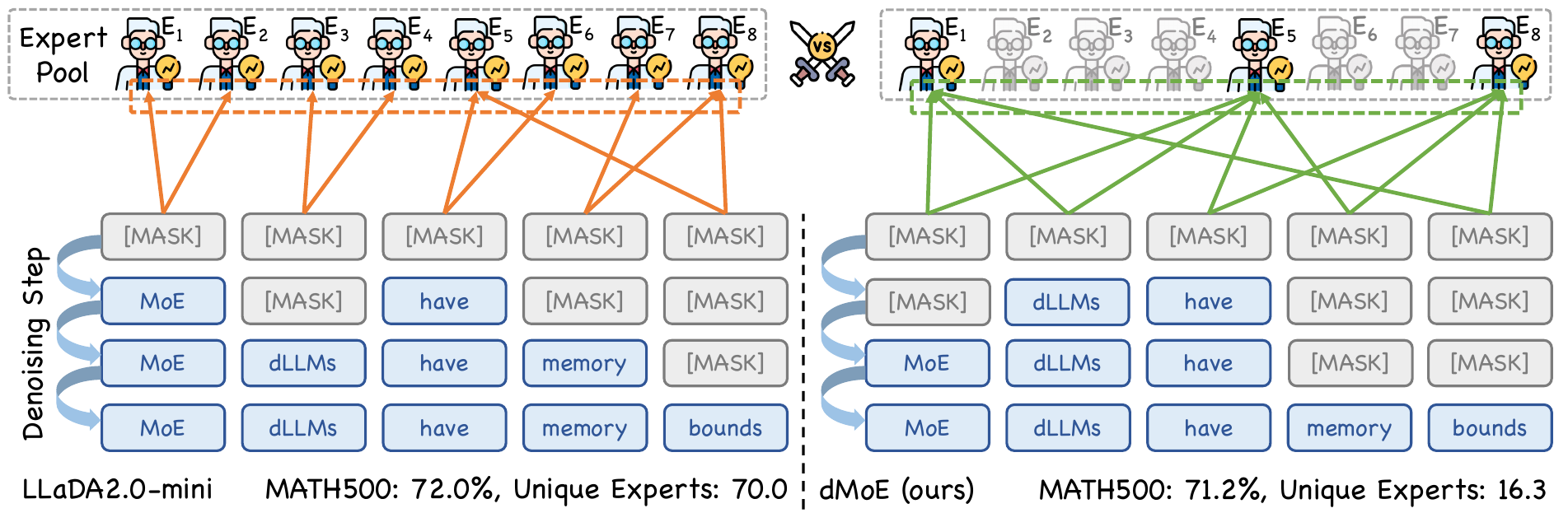} \\ %
\vspace{-4mm}
\caption{Comparison between the original LLaDA2.0-mini and our proposed \method. Unlike the original token-level expert routing in MoE dLLMs, our \method replaces token-level routing with block-level routing, substantially reducing the unique expert count while preserving performance.}
\vspace{-4mm}
\label{fig:intro-comparison}
\end{figure*}

\section{Introduction}

Recently, diffusion large language models (dLLMs)~\citep{yi2024diffusion,zhang2025survey,nie2025large,ye2025dream,yu2025dimple} have emerged as a competitive alternative to autoregressive LLMs~\citep{achiam2023gpt,bai2023qwen,dubey2024llama}, demonstrating strong performance in both open-source and closed-source settings~\citep{song2025seed,khanna2025mercury}. By iteratively refining mask tokens through an unmask-and-remask process, dLLMs naturally support parallel decoding beyond the left-to-right generation order of autoregressive models, offering greater flexibility and efficiency potential at test time. To further scale model capacity while keeping the number of active parameters manageable, recent dLLMs~\citep{bie2025llada2,bie2026llada21speedingtextdiffusion,zhu2025lladamoe,cheng2025sdar,ni2025openmoe2,lin2026efficient} have increasingly been integrated with Mixture-of-Experts (MoE) architectures, making MoE a growing design trend in this paradigm.

While MoE architectures offer an effective scaling strategy by increasing model capacity with sparse activation, they also introduce a fundamental efficiency challenge. In MoE dLLMs, existing MoE routing still follows the token-level expert selection paradigm inherited from autoregressive models, selecting experts independently for each token. However, dLLMs process multiple tokens within a single forward pass (\textit{e.g.}, a whole block of tokens in block diffusion decoding~\citep{arriola2025block}). As a result, the number of uniquely activated experts can grow dramatically within one forward pass, making memory access the primary inference bottleneck. The empirical results in Section~\ref{sec:preliminaries} further support our claims by showing that MoE latency dominates the end-to-end inference latency and linearly increases with the number of uniquely activated experts.

A large body of prior work has studied efficient MoE strategies, primarily in autoregressive models. 
Existing methods can be broadly divided into two categories: pre-execution compression, such as expert pruning~\citep{liu2024efficient,chen2022task,chowdhury2024provably,guo2025cluster,song2025blockffn} and expert merging~\citep{he2023merging,park2024learning,li2026sub}, and runtime-adaptive execution, such as expert skipping~\citep{lu2024not,huang2025modes,aghdam2024moe}, adaptive expert selection~\citep{chen2025eac}, and expert reuse~\citep{tan2025rexmoe,oncescu2025opportunistic}.
However, MoE efficiency in dLLMs remains largely underexplored. Unlike autoregressive decoding, dLLMs perform parallel token generation and refinement within each denoising step, making expert activation patterns and efficiency bottlenecks fundamentally different. A few recent studies have begun to explore this setting. EC-DLM~\citep{zhang2026expert} replaces token-choice routing with expert-choice routing and improves load balancing by dynamically adjusting expert capacity. TEAM~\citep{wei2026team} leverages temporal and spatial consistency in routing to reuse experts across denoising steps, thereby reducing overall expert activation during inference. DES~\citep{chen2026dynamic} explicitly targets the memory overhead caused by excessive expert activation through a candidate-constrained routing strategy. 

We propose \method, a simple yet effective strategy for compressing unique experts in MoE dLLMs. Our design is motivated by two key observations. First, token-level expert scores provide an informative signal of expert importance. Second, the degree of expert concentration varies substantially across denoising steps and blocks.
Specifically, we first aggregate token-level expert scores to form block-level expert scores, and then use these block-level scores to guide the original routing process, thereby controlling the number of uniquely activated experts. 
In this way, \method can aggressively reduce the unique expert count without changing the number of experts selected for each token. 
Moreover, \method dynamically controls the unique expert count with a top-p criterion, allowing it to better adapt to the varying routing characteristics across different denoising steps and blocks. During training, we adopt a self-distillation paradigm using the same routing procedure in the forward pass.

We choose LLaDA2.0-mini, a popular open-source dLLM, as the base model for fine-tuning and evaluation. We evaluate \method on six benchmarks, including MATH500~\citep{lightman2023let}, GSM8K~\citep{cobbe2021gsm8k}, ARC-C~\citep{clark2018think}, MMLU~\citep{hendryckstest2021}, TruthfulQA~\citep{lin2022truthfulqa}, and QASPER~\citep{dasigi2021dataset}. The results show that \method consistently achieves substantial expert compression with no performance degradation (Figure~\ref{fig:intro-comparison}). 
On average, \method reduces the number of uniquely activated experts by \(4.19\times\) while retaining 99.55\% of the original performance. In addition, it reduces memory usage by 74.36\% to 76.78\% and delivers \(1.36\times\) to \(2.04\times\) end-to-end latency speedup compared with the original model. Our \method also achieves a superior performance-efficiency trade-off compared with the baselines. Furthermore, our \method is tunable, allowing the number of activated experts to be adjusted to different application requirements.

Overall, we introduce \method, a novel learnable strategy for block-level expert routing in MoE dLLMs. The core idea of \method is to aggregate token-level expert scores into block-level expert scores, and then use these block-level scores to dynamically guide the original routing process. Extensive experiments demonstrate the effectiveness of our method. This work establishes a strong baseline for block-level routing in MoE dLLMs.

\section{Related Work}
\label{sec:related-works}

\noindent \textbf{Overview of Diffusion Language Models.} 
Diffusion-based generative modeling has achieved remarkable success in continuous modalities, including images~\citep{rombach2022high,peebles2023scalable}, videos~\citep{ho2022video,brooks2024video}, and audio~\citep{liu2023audioldm,evans2024fast}, building on the broader foundation of diffusion models~\citep{ho2020denoising,song2019generative,song2020denoising}. Extending this framework to language, however, is non-trivial because text is inherently discrete. To address this challenge, a growing body of work formulates diffusion directly in token space~\citep{austin2021structured,sahoo2024simple,lou2023discrete,zheng2024masked,cheng2025sdar,nie2025large,chen2025dparallel,ma2025dkv,2026QiuhongDLLM}, often through masked-token denoising, which enables parallel generation and relaxes the strict left-to-right dependency of autoregressive decoding. Based on this formulation, dLLMs~\citep{nie2025large,ye2025dream,khanna2025mercury,song2025seed,bie2025llada2} have shown increasingly competitive performance at the billion-parameter scale, suggesting that diffusion is becoming a practical alternative for language generation. 
Recent progress also suggests a clear trend toward sparse scaling in dLLMs, with an increasing number of representative models adopting MoE architectures to expand overall model capacity while keeping the number of active parameters at each denoising step relatively small~\citep{zhu2025lladamoe,bie2025llada2,bie2026llada21speedingtextdiffusion}.
Beyond general text generation, diffusion-based language modeling is now being extended to more challenging settings, including reasoning~\citep{zhu2025llada,zhao2025d1,tang2025wd1,lin2025boundary,feng2026dvoting,feng2025efficient}, multimodal generation~\citep{yang2025mmada,li2025lavida,yu2025dimple,you2025llada}, and code synthesis~\citep{gong2025diffucoder,khanna2025mercury,pengcontributors}, highlighting the rapid expansion and growing maturity of this research direction~\citep{yu2025discrete,li2025survey,chen2026dmax}.

\noindent \textbf{Efficient Mixture-of-Expert Strategies.}
In autoregressive models, most existing studies in this setting are designed to reduce token-wise expert computation or serving overhead, and are therefore primarily tailored to left-to-right decoding. Broadly, these methods can be grouped into two lines. The first line focuses on pre-execution compression, which reduces the model-side budget before inference, for example, through expert pruning~\citep{liu2024efficient,chen2022task,chowdhury2024provably,guo2025cluster,song2025blockffn} and expert merging~\citep{he2023merging,park2024learning,li2026sub}. The second line focuses on runtime-adaptive execution, which improves efficiency during inference by dynamically controlling expert activation according to the current input, such as expert skipping~\citep{lu2024not,huang2025modes,aghdam2024moe}, adaptive expert selection~\citep{chen2025eac}, and expert reuse~\citep{tan2025rexmoe,oncescu2025opportunistic}.
Beyond these algorithmic strategies, another important direction lies in system-level optimization~\citep{sarkar2023edge,he2022fastermoe}, which improves MoE efficiency from the perspectives of communication and memory.

\noindent However, MoE efficiency in dLLMs remains largely underexplored. Some recent studies have provided initial evidence showing its potential. EC-DLM~\citep{zhang2026expert} replaces conventional token-choice routing with expert-choice routing and further improves load balancing by dynamically adjusting expert capacity across denoising steps. TEAM~\citep{wei2026team} exploits temporal and spatial consistency in expert routing to reuse experts across denoising steps, thereby reducing the overall number of activated experts during inference. DES~\citep{chen2026dynamic} explicitly targets the memory overhead caused by excessive expert activation within each block under parallel decoding. It introduces a candidate-constrained routing strategy to compress the number of block-level activated experts without any additional training. 
Our target is closely aligned with DES, but pushes this direction further by pursuing more aggressive block-level expert selection to achieve stronger expert compression.

\section{Preliminaries}
\label{sec:preliminaries}

\begin{figure*}[t]
\centering
\includegraphics[width=\linewidth]{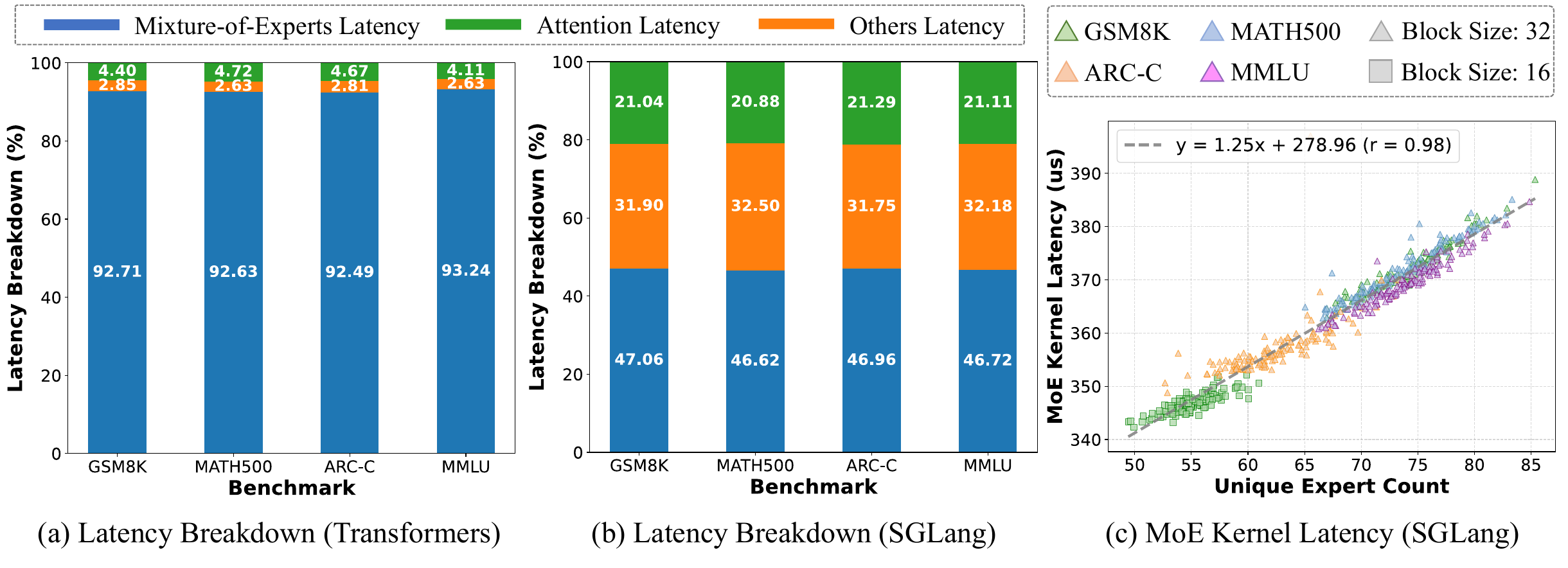}  \\
\vspace{-4mm}
\caption{Empirical studies on LLaDA2.0-mini. (a) \& (b) We report the latency breakdown of three components: MoE (\textit{e.g.}, routing and execution), attention (\textit{e.g.}, attention computation and related projections), and others (\textit{e.g.}, LM head and embeddings). (c) We present the relationship between the number of unique activated experts and MoE kernel latency.} %
\label{fig:preliminaries}
\vspace{-3mm}
\end{figure*}

We briefly introduce dLLMs from the perspective most relevant to this work: their native support for parallel decoding. Unlike autoregressive language models, which generate tokens strictly from left to right, dLLMs predict multiple unknown positions simultaneously by iteratively refining a partially masked sequence. This property makes them especially suitable for parallel generation, since the model can update a set of masked tokens in one denoising step rather than committing to a single next token at a time. A common instantiation is the masked diffusion language model (MDLM), where generation starts from a corrupted sequence $x_t$ obtained by masking tokens from a clean sequence $x_0$. Let $t \in [0,1]$ denote the masking intensity. The forward corruption process independently replaces each token with \text{[MASK]} according to
\begin{equation}
q(x_t \mid x_0)
=
\prod_{i=1}^{L}
\left[
(1-t)\,\delta(x_t^{i}=x_0^{i})
+
t\,\delta(x_t^{i}=\text{[MASK]})
\right].
\end{equation}
Given the corrupted sequence, the denoising model $p_\theta$ predicts the original tokens at the masked positions:
\begin{equation}
p_\theta(x_0 \mid x_t)
=
\prod_{i: x_t^i=\text{[MASK]}}
p_\theta(x_0^i \mid x_t).
\end{equation}
Because all masked positions can be reconstructed in parallel within each denoising step, dLLMs naturally enable non-autoregressive parallel decoding.

\noindent \textbf{Memory Bound Becomes the Primary Bottleneck for Mixture-of-Experts dLLMs.}
In MoE dLLMs, each forward pass needs to select experts for multiple tokens simultaneously (\textit{e.g.}, a full block of tokens under block diffusion decoding~\citep{arriola2025block}). As a result, a large number of unique experts can be activated within a single forward pass, and these experts need to be iteratively loaded during inference, leading to substantial memory overhead. We further conduct empirical studies on LLaDA2.0-mini across various benchmarks to support this claim. Specifically, for all empirical studies, we use block diffusion decoding and include a warm-up stage before measurement to stabilize memory, cache, and kernel states. 
As shown in Figure~\ref{fig:preliminaries}~(a) and (b), we report the latency breakdown of three components: MoE, attention, and others. Here, MoE includes routing and expert execution, attention includes attention computation and related projections, and others include components such as the LM head and embeddings. We report results under both the native Transformers\footnote{\url{https://github.com/huggingface/transformers}} and SGLang\footnote{\url{https://github.com/sgl-project/sglang}} frameworks, and consistently observe that MoE latency dominates the end-to-end latency. In Figure~\ref{fig:preliminaries}~(c), we additionally record the MoE kernel latency together with the corresponding unique expert count, \textit{i.e.}, the number of uniquely activated experts within a single forward pass; under block diffusion, this corresponds to the unique experts activated by all tokens in the current block. The results show a clear linear positive correlation, indicating that activating more unique experts leads to higher MoE kernel latency. Overall, these findings motivate us to alleviate the memory bottleneck in MoE dLLMs by reducing the number of unique activated experts.

\begin{figure*}[t]
\centering
\resizebox{0.995\linewidth}{!}{
\begin{tabular}{cc} %
  \includegraphics[width=0.48\linewidth]{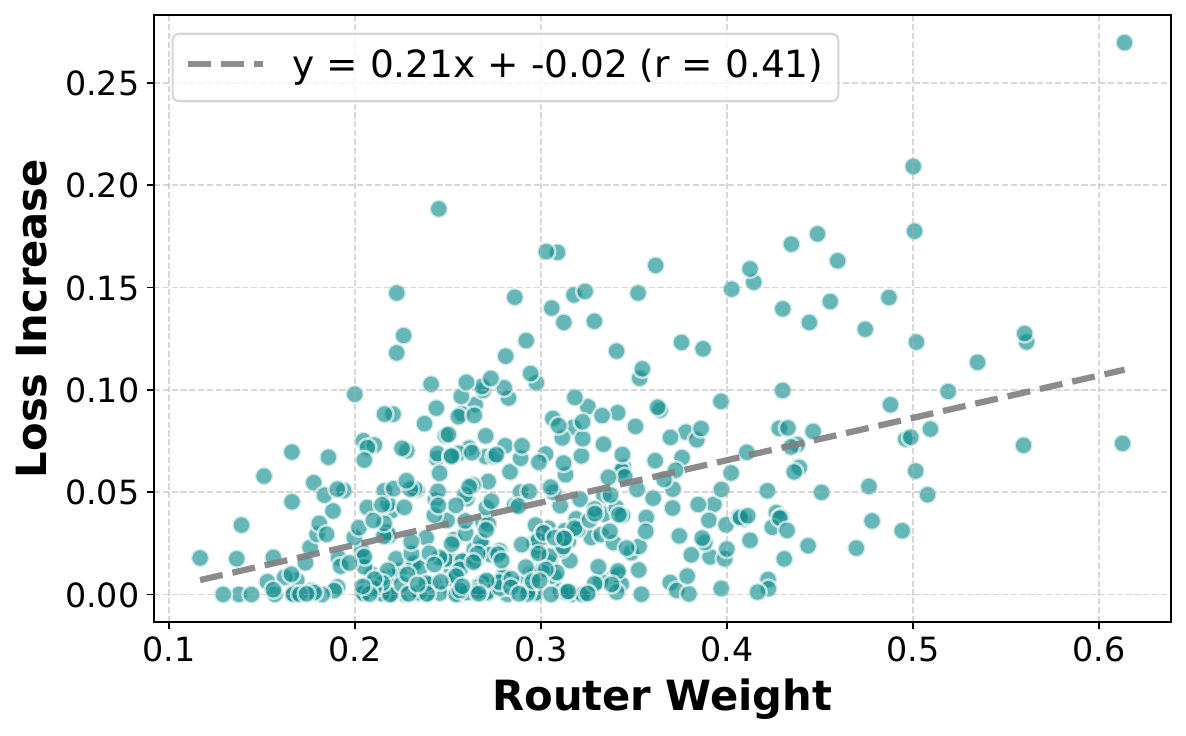} &
  \includegraphics[width=0.48\linewidth]{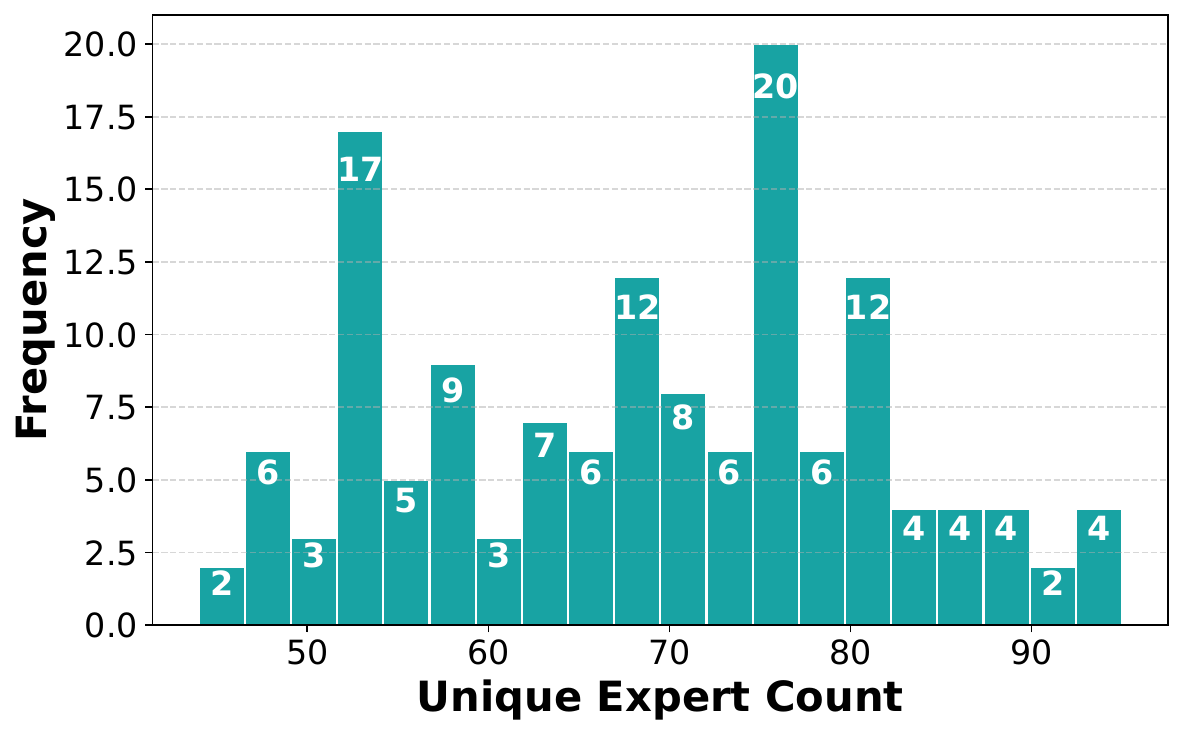} \\
  {\small (a) Router weight \textit{vs.} Loss increase} & {\small (b) Distribution of unique expert count} \\
\end{tabular}}
\vspace{-3mm}
\caption{(a) Router weights of the masked experts (\(x\)-axis) \textit{vs.} the resulting loss increase (\(y\)-axis) at Layer 10 (the middle layer), with 432 data points sampled across denoising steps on GSM8K. (b)~Distribution of the unique expert count during inference on GSM8K (details in Appendix~\ref{apx:expert-count}).}
\label{fig:observations}
\end{figure*}

\section{Methods}
\label{sec:method}

\begin{figure*}[t]
\centering
  \includegraphics[width=\linewidth]{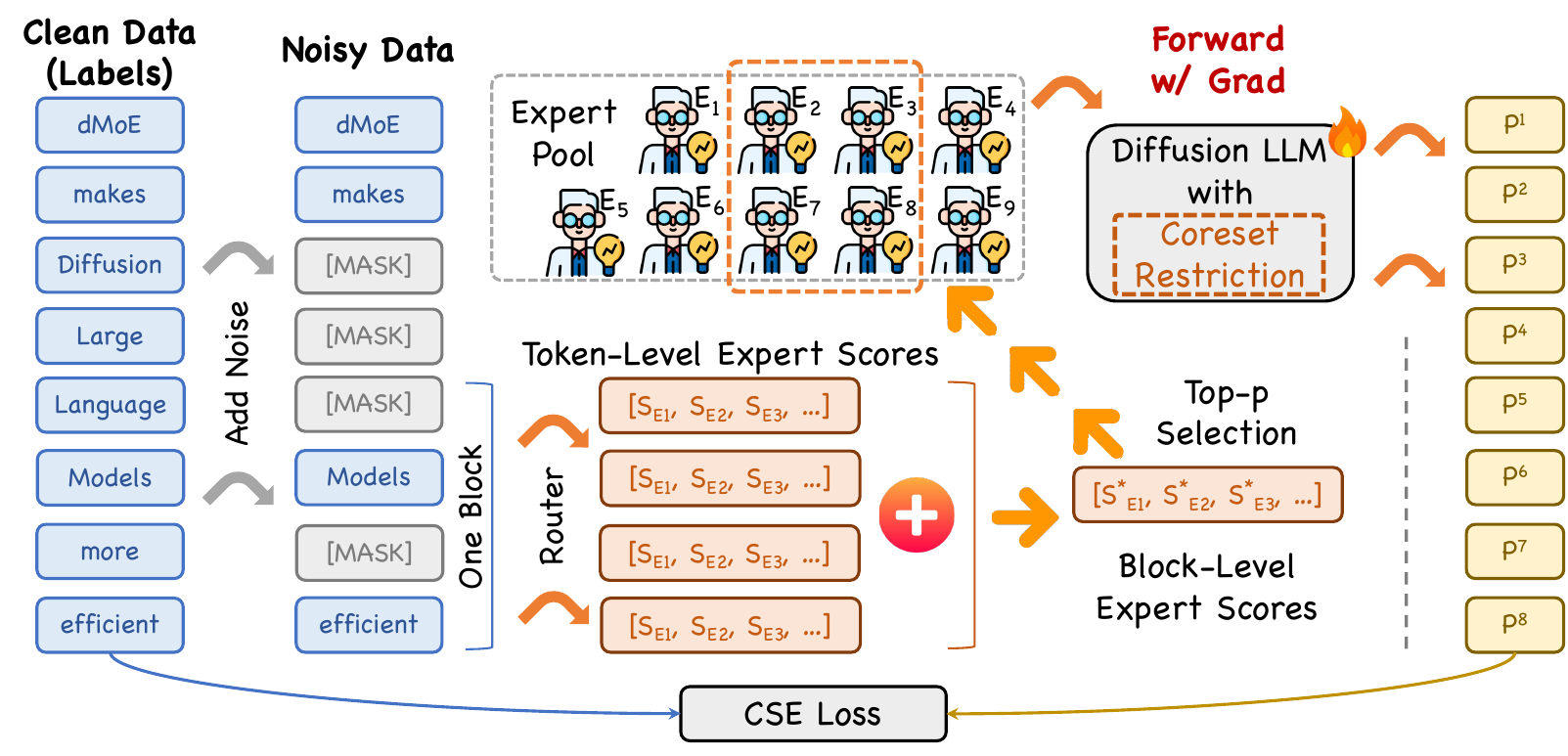} \\ %
\vspace{-4mm}
\caption{Overview of our proposed \method. For each noisy block, we aggregate token-level router scores into block-level expert scores and apply a top-p criterion to select an adaptive expert coreset. The final token-level routing is then restricted to this coreset. Training follows a self-distillation paradigm using CSE loss, and inference uses the same forward process.}
\label{fig:overview}
\vspace{-2mm}
\end{figure*}

Based on the analysis in Section~\ref{sec:preliminaries}, we focus on reducing the number of block-level unique experts to alleviate the memory bottleneck while keeping the computation unchanged, that is, maintaining a fixed number of selected experts for each token. In this section, we first present two key observations in Section~\ref{subsec:observations}, and then introduce our proposed \method in detail in Section~\ref{subsec:dslim}.

\subsection{Key Observations}
\label{subsec:observations}

We conduct empirical studies of block diffusion decoding on LLaDA2.0-mini. From these results, we identify several key observations that motivate the design of our \method for expert compression in MoE dLLMs.

\noindent \textbf{Observation A.}
\textit{Token-level expert scores provide an effective indicator of expert importance.} To verify this, we analyze the relationship between token-level expert scores (\textit{i.e.}, router weights) and the change in reconstruction loss after masking the corresponding experts. Specifically, we randomly sample 432 data points from GSM8K, using the original LLaDA2.0-mini as the base model, and compute the correlation between these two quantities. As shown in Figure~\ref{fig:observations}~(a), they exhibit a clear positive correlation, with a Pearson correlation coefficient of 0.41. This observation motivates us to aggregate token-level expert scores for block-level expert selection.

\begin{wraptable}{t}{0.52\columnwidth}
\centering
\vspace{-5mm}
\caption{Results of the coefficient of variation (CV) on various benchmarks.}
\label{tab:observation-b}
\vspace{1mm}
\setlength{\tabcolsep}{1.5mm}
\resizebox{\linewidth}{!}{
\begin{tabular}{lcccc}
\toprule
\textbf{Metric} 
& \textbf{GSM8K} 
& \textbf{MATH500} 
& \textbf{ARC-C}
& \textbf{MMLU} \\
\midrule
Mean & 70.01 & 71.09 & 62.85  & 66.57 \\
Std & 11.96 & 11.34 & 10.01 & 11.85 \\
CV & 17.09\% & 15.95\% & 15.93\% & 17.80\% \\
\bottomrule
\end{tabular}}
\vspace{-5mm}
\end{wraptable}

\noindent \textbf{Observation B.}
\textit{The degree of expert concentration varies substantially across denoising steps and blocks.} 
To examine this phenomenon in more detail, we further analyze the distribution of the unique expert count for the generated block at Layer 10 during inference, where Layer 10 corresponds to the middle layer of LLaDA2.0-mini, which contains 19 MoE layers in total. As illustrated in Figure~\ref{fig:observations}~(b), the unique expert count changes noticeably across different denoising steps, suggesting that the extent to which tokens are routed to a concentrated subset of experts is highly step-dependent. To further quantify this variability, we adopt the coefficient of variation (CV), defined as
\[
\mathrm{CV} = \frac{\sigma}{\mu},
\]
where \(\sigma\) and \(\mu\) denote the standard deviation and mean of the unique expert count, respectively. As shown in Table~\ref{tab:observation-b}, the unique expert count exhibits consistently large variation across various benchmarks, suggesting that this phenomenon is not tied to a specific dataset or task. These results provide further empirical evidence that expert concentration varies substantially across denoising steps and blocks.

\subsection{\method}
\label{subsec:dslim}

To achieve our goal of reducing the number of block-level unique experts, a natural idea is to constrain the expert pool during expert selection. Following a coarse-to-fine strategy, we first select a coreset from the full expert pool, and then perform token-level expert selection within this coreset (Figure~\ref{fig:overview}). Specifically, given the tokens within a block, we first perform token-level routing to compute token-level expert scores:
\[
\mathbf{s}_i = \operatorname{Router}(t_i)
= \left[s_{E_1},\, s_{E_2},\, \dots,\, s_{E_{|\mathcal{E}|}}\right],
\]
where \(t_i\) denotes the representation of the \(i\)-th token in the block, and \(s_{E_i}\) is the routing score of expert \(E_i\) for token \(i\). Inspired by observation A, which suggests that token-level expert scores are positively correlated with expert importance, we then directly aggregate the token-level expert scores to obtain block-level expert scores:
\[
S_{block} = \oplus_{i \in \mathcal{B}} s_{i},
\]
where \(\mathcal{B}\) denotes the set of tokens in the current block, and \(\oplus\) represents the aggregation operator. After obtaining the block-level expert scores, we further determine the coreset based on their normalized values. Inspired by observation B, which suggests that the degree of expert concentration varies substantially across denoising steps and blocks, we do not enforce a fixed coreset size. Instead, we first normalize the block-level expert scores and then apply a top-p criterion to select the coreset:
\[
\mathcal{C} = \operatorname{top-p}(\{\tilde{S}_{block}\}_{e=1}^{E},\, p),
\]

\begin{wrapfigure}{r}{0.52\columnwidth}
    \vspace{-1.5em}
    \begin{minipage}{0.52\columnwidth}
    \small
    \begin{algorithm}[H]
    \caption{Block-Level Expert Routing}
    \label{alg:block_expert_routing}
    \KwIn{Token-level expert scores \(\{\mathbf{s}_i\}_{i\in\mathcal{B}}\)}
    \KwOut{Block-level expert scores \(S_{block}\), coreset \(\mathcal{C}\), routed experts \(\{R_i\}_{i\in\mathcal{B}}\)}

    Initialize \(S_{block} \leftarrow \mathbf{0} \in \mathbb{R}^{|\mathcal{E}|}\)\;
    \ForEach{\(i \in \mathcal{B}\)}{
        \(\hat{\mathbf{s}}_i \leftarrow \operatorname{TopKMask}(\mathbf{s}_i, k)\)\;
        \(S_{block} \leftarrow S_{block} + \hat{\mathbf{s}}_i\)\;
    }

    \(\tilde{S}_{block} \leftarrow \operatorname{Normalize}(S_{block})\)\;
    \(\mathcal{C} \leftarrow \operatorname{top-p}(\tilde{S}_{block}, p)\)\;

    \ForEach{\(i \in \mathcal{B}\)}{
        \(\mathbf{s}_i^{\mathcal{C}} \leftarrow \operatorname{MaskOutside}(\mathbf{s}_i, \mathcal{C})\)\;
        \(R_i \leftarrow \operatorname{TopK}(\mathbf{s}_i^{\mathcal{C}}, k)\)\;
    }

    \Return \(S_{block}, \mathcal{C}, \{R_i\}_{i\in\mathcal{B}}\)\;
    \end{algorithm}
    \end{minipage}
    \vspace{-2em}
\end{wrapfigure}

where \(\mathcal{C}\) denotes the selected coreset and \(p\) is the cumulative probability threshold. This design naturally yields a smaller coreset when the expert score distribution is concentrated, and a larger one when the distribution is more dispersed, thereby adapting to the routing characteristics of different blocks. As such, it is better aligned with the original inference behavior. Finally, the original token-level routing is performed within the selected coreset (see more details in Algorithm~\ref{alg:block_expert_routing}).

During the training process, we adopt a self-distillation paradigm~\cite{zhang2021self} and follow the above routing procedure in the forward pass. During the inference process, we use the same forward process to maintain alignment between training and inference.

\section{Experiments}
\label{sec:experiments}

\subsection{Experimental Setup}
\label{sec:experimental-setup}

\noindent \textbf{Training Data.}
All training data are constructed via self-distillation~\citep{zhang2021self}. We first collect prompts from several public datasets, including the GSM8K training set~\citep{cobbe2021gsm8k}, PRM12K~\citep{lightman2023let}, a subset of Numina-Math~\citep{li2024numinamath}, and a subset of OpenThoughts~\citep{guha2025openthoughts}. We then use LLaDA2.0-mini (\textit{i.e.}, our base model) to generate corresponding responses as supervision targets. During generation, we follow the official settings: the confidence threshold is set to 0.95, the block size to 32, and the maximum output length to 2048 tokens. Samples that fail to terminate within this length limit are removed. Importantly, the entire supervision signal is derived from the model’s own generations, without relying on any externally curated high-quality responses. Overall, we obtain approximately 700K training samples.

\noindent \textbf{Training Details.}
We use LLaDA2.0-mini as the base model, which is a popular open-source MoE dLLM. During training, we randomly sample the masking ratio from \([0.3, 0.8]\) for each training instance. We perform full fine-tuning for 2 epochs with a global batch size of 4. The learning rate is set to \(2.0 \times 10^{-6}\) with a cosine learning rate schedule. We adopt the block diffusion setting and follow the official configuration by setting the block size to 32. We train two models with the top-p threshold \(p^{train}\) set to 0.6 (default) and 0.5, respectively. All training is conducted on 4 H100 GPUs.

\noindent \textbf{Baselines.}
We compare \method against six baselines: (1) \textbf{Original}, which follows the official configuration with block diffusion inference; (2) \textbf{+ Fine-Tuning}, which keeps the original routing but applies the same self-distillation fine-tuning as \method; (3) \textbf{Top-4}, which reduces the number of selected experts per token from 8 in the official setting to 4; (4) \textbf{DES-S}, where we implement DES-Seq~\citep{chen2026dynamic}, a sequence-level routing strategy that enables more adaptive expert allocation; (5) \textbf{DES-V}, where we implement DES-Vote~\citep{chen2026dynamic}, which reduces the number of unique experts within each block by consolidating router-derived preferences across tokens; and (6) \textbf{TEAM}~\citep{wei2026team}, which reuses experts across denoising steps by exploiting the temporal and spatial consistency in expert routing. Additionally, for both DES-S and DES-V, the degree of expert compression can be controlled through hyperparameters.

\noindent \textbf{Inference Details.}
We follow the official inference settings to evaluate the LLaDA2.0-mini~\citep{bie2025llada2}. For the baselines, we report the results from our implementation, with the latter following the configurations specified in the original paper.
We apply the block diffusion setting with confidence-based parallel decoding, consistent with the official implementation. We follow the official settings for both our \method and all the baselines: set block size to 32, set max generation length to 2048, set confidence threshold to 0.95, and enable early stopping. All results in this paper are obtained with KV caching enabled for all methods. Unless otherwise specified, we use a batch size of 1 for inference, and results with larger batch sizes are provided in Table~\ref{tab:batch-size}.

\noindent \textbf{Evaluation Details.}
We conduct comprehensive experiments on multiple benchmarks spanning a broad spectrum of reasoning tasks, including mathematical reasoning, scientific reasoning, and general high-level reasoning. Specifically, we evaluate on MATH500~\citep{lightman2023let}, GSM8K~\citep{cobbe2021gsm8k}, ARC-C~\citep{clark2018think}, and MMLU~\citep{hendryckstest2021}. In addition, we follow the simple-eval framework\footnote{\url{https://github.com/openai/simple-evals}} for zero-shot evaluation and prompt the model to generate its reasoning trajectories step by step. Beyond reasoning tasks, we further evaluate on two general benchmarks, TruthfulQA~\citep{lin2022truthfulqa} and QASPER~\citep{dasigi2021dataset}.

\subsection{Main Results}
\label{sec:main-results}

\begin{table}[t]
\centering
\caption{Results on various benchmarks with LLaDA2.0-mini. For MMLU, we use the math part for evaluation. Since the official DES code is not publicly available, we implement DES for comparison. Expert count denotes the average number of unique activated experts per layer. ${*}$ indicates that the corresponding threshold in DES is adjusted to achieve a higher compression ratio.}
\label{tab:llada2-main-1}
\vspace{-1.5mm}
\resizebox{0.995\linewidth}{!}{
\setlength{\tabcolsep}{1mm}
\begin{tabular}{lcccccccc}
\toprule
\multirow{2}{*}{\textbf{Method}} 
& \multicolumn{2}{c}{\textbf{MATH500}} 
& \multicolumn{2}{c}{\textbf{GSM8K}}
& \multicolumn{2}{c}{\textbf{ARC-C}}
& \multicolumn{2}{c}{\textbf{MMLU}} \\ 
\cmidrule(l{4pt}r{4pt}){2-3} \cmidrule(l{4pt}r{4pt}){4-5} \cmidrule(l{4pt}r{4pt}){6-7} \cmidrule(l{4pt}r{4pt}){8-9}
&  Expert Count & Accuracy 
&  Expert Count & Accuracy 
&  Expert Count & Accuracy 
&  Expert Count & Accuracy \\ 
\midrule%
LLaDA2.0-mini & 
70.0 & 72.0\%  & 
71.5 & 90.8\% & 
65.9 & 84.4\%  &
70.6 & 88.6\% \\ 
\midrule
\addlinespace
\multicolumn{9}{c}{\textit{Baselines}} \\
\midrule
+ Fine-Tuning &
79.1 & 72.2\%  &
79.7 & 91.0\%  &
70.7 & 87.0\%  &
78.9 & 89.2\%  \\
Top-4 &
43.5 & 52.6\%  & 
44.3 & 72.4\% & 
42.2 & 74.0\%  &
43.8 & 61.5\%  \\
DES-S  &
41.0 & 70.8\%  &
41.9 & 90.7\%  &
39.9 & 83.3\%  &
41.4 & 87.1\%  \\
DES-V  &
36.9 & 71.8\%  &
37.0 & 90.5\%  &
36.8 & 82.3\%  &
37.0 & 85.6\%  \\
DES-S$^{*}$  &
12.9 & 53.8\%  &
13.2 & 74.0\%  &
13.0 & 62.9\%  &
13.0 & 70.7\%  \\
DES-V$^{*}$ &
13.0 & 53.0\%  &
12.9 & 73.5\%  &
13.0 & 49.1\%  &
13.0 & 72.6\%  \\
TEAM &
43.6 & 71.2\%  &
44.8 & 90.1\%  &
38.4 & 85.0\%  &
43.8 & 87.4\%  \\
\midrule
\addlinespace
\multicolumn{9}{c}{\method} \\
\midrule
\rowcolor{gray!30}  ours (\(p^{train} = p^{test} = 0.6\))  &
16.3 & 71.2\%  &
16.6 & 88.6\%  &
16.9 & 87.1\%  &
16.5 & 87.4\%  \\
\rowcolor{gray!30}  ours (\(p^{train} = p^{test} = 0.5\))  &
13.7 & 69.8\%  &
13.8 & 87.1\%  &
13.9 & 84.7\%  &
13.6 & 86.7\%  \\
\bottomrule
\end{tabular}}
\end{table}

\begin{figure*}[t]
\centering
\resizebox{0.995\linewidth}{!}{
\begin{tabular}{cc} %
  \includegraphics[width=0.48\linewidth]{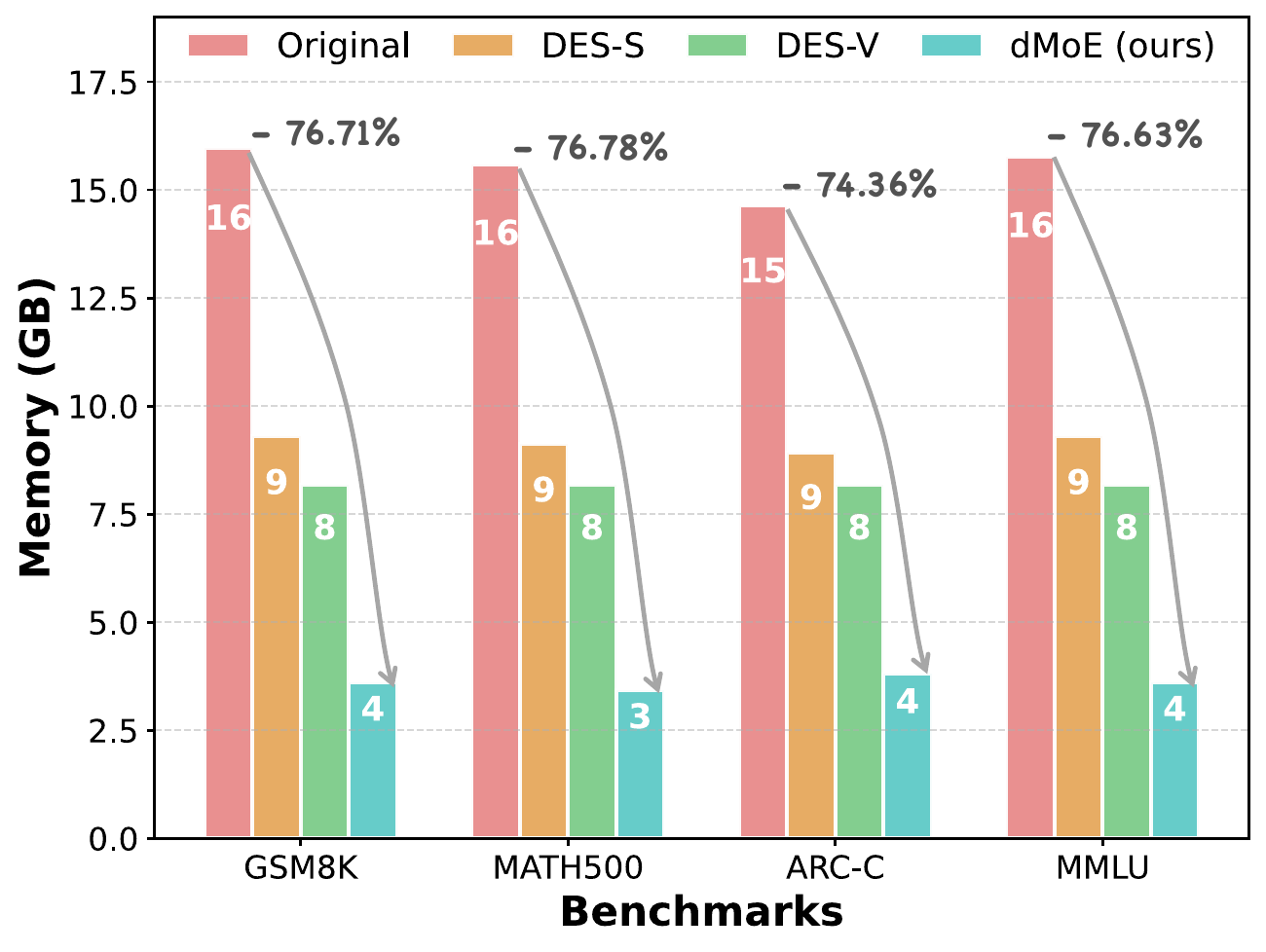} &
  \includegraphics[width=0.48\linewidth]{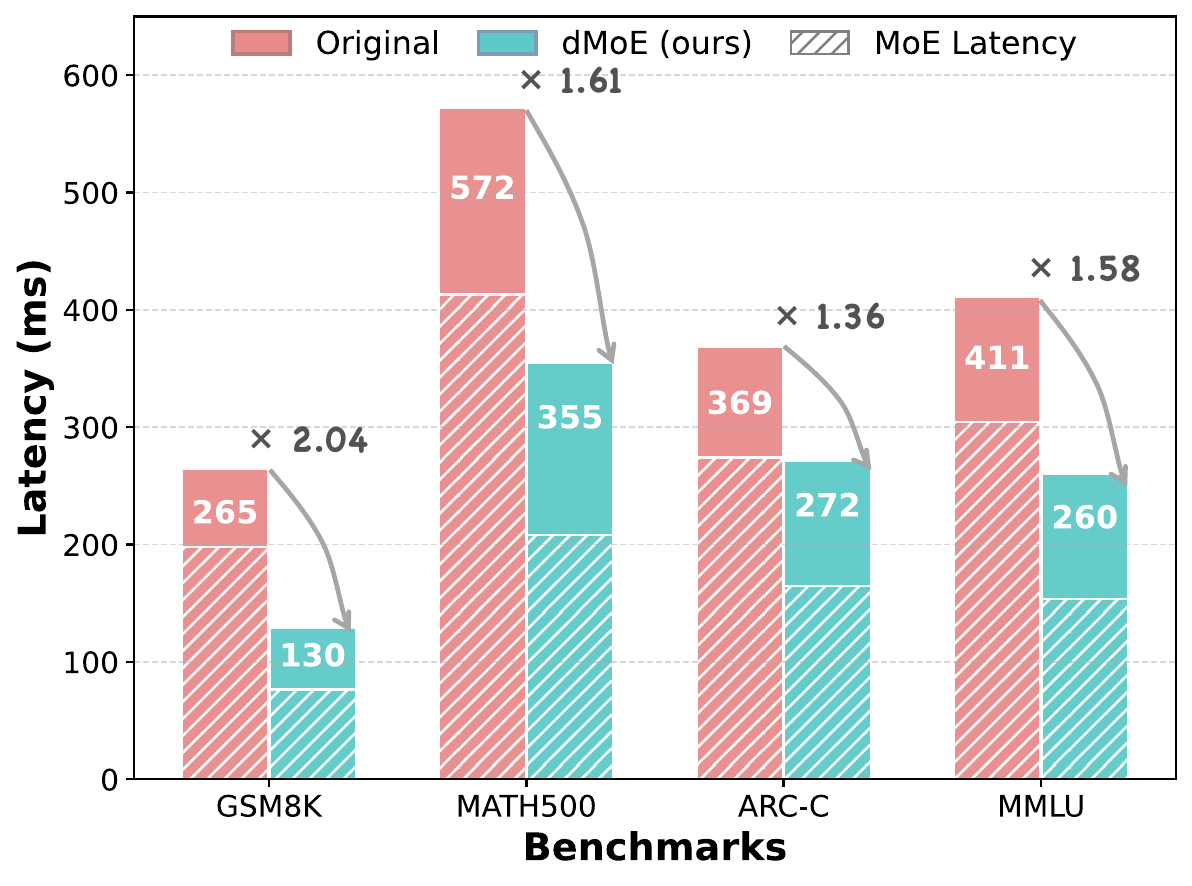} \\
  {\small (a) Memory comparison} & {\small (b) Latency speedup} \\
\end{tabular}}
\vspace{-3mm}
\caption{(a) We report the average memory footprint of uniquely activated MoE parameters across all layers for both the baselines and our \method. (b) We present the average end-to-end latency and MoE latency for the original model and our \method.} %
\label{fig:memory-latency}
\vspace{-4mm}
\end{figure*}

\noindent \textbf{\method Achieves Aggressive Expert Compression While Preserving Performance.}
We report the main results in Table~\ref{tab:llada2-main-1}. Compared with the original model, our \method achieves substantial expert compression while preserving performance almost intact. Specifically, the average performance across four benchmarks decreases only slightly from 83.95\% to 83.575\%, retaining 99.55\% of the original performance, while reducing the average unique expert count from 69.5 to 16.575, corresponding to a 76.15\% reduction in the average unique expert count. Compared with the baselines, our \method further achieves a 55.11\% to 61.14\% greater reduction in unique expert count at a comparable performance level. Moreover, under the same self-distillation fine-tuning, \method retains 98.50\% of the performance of + Fine-Tuning with 78.50\% fewer unique experts, indicating that the expert compression stems from block-level routing rather than fine-tuning.

\noindent \textbf{\method Generalizes Beyond Reasoning Tasks.}
We further report the results on two general benchmarks in Table~\ref{tab:general}. \method reduces the average unique expert count from 61.25 to 15.4, corresponding to a 74.86\% reduction, while maintaining comparable overall performance. Notably, \method improves the accuracy on TruthfulQA from 62.2\% to 69.9\%, with only a slight drop on QASPER from 66.5\% to 64.8\%. These results demonstrate that the effectiveness of \method extends beyond reasoning tasks.

\begin{wraptable}{r}{0.51\columnwidth}
\vspace{-1.5em}
\centering
\caption{Results on general benchmarks. For TruthfulQA, we report BLEU accuracy. EC denotes the average number of unique activated experts.} %
\label{tab:general}
\vspace{1mm}
\setlength{\tabcolsep}{1.5mm}
\resizebox{\linewidth}{!}{
\begin{tabular}{lcccc}
\toprule
\multirow{2}{*}{\textbf{Method}}
& \multicolumn{2}{c}{\textbf{TruthfulQA}}
& \multicolumn{2}{c}{\textbf{QASPER}} \\
\cmidrule(l{4pt}r{4pt}){2-3} \cmidrule(l{4pt}r{4pt}){4-5}
&  \begin{tabular}[b]{@{}c@{}}EC\end{tabular} & Accuracy
&  \begin{tabular}[b]{@{}c@{}}EC\end{tabular} & Accuracy \\
\midrule
LLaDA2.0-mini &
61.2 & 62.2\%  &
61.3 & 66.5\%  \\
\rowcolor{gray!30}  ours  &
15.4 & 69.9\%  &
15.4 & 64.8\%  \\
\bottomrule
\end{tabular}}
\vspace{-1.5em}
\end{wraptable}

\noindent \textbf{\method Significantly Reduces Memory Usage.}
As shown in Figure~\ref{fig:memory-latency} (a), we compare the memory footprint of the original model, the baselines, and our \method. Specifically, we report the average memory footprint of uniquely activated MoE parameters across all layers. The results show that our \method achieves the lowest memory usage, yielding a 74.36\% to 76.78\% reduction compared with the original model and a 54.08\% to 60.38\% reduction compared with the baselines.

\begin{table}[t]
\centering
\vspace{1mm}
\caption{Latency comparison with DES on LLaDA2.0-mini. We report the average end-to-end latency (ms) and the speedup over the original model. All methods are compared at comparable accuracy.}
\label{tab:latency}
\vspace{-1.5mm}
\resizebox{0.995\linewidth}{!}{
\setlength{\tabcolsep}{1mm}
\begin{tabular}{lcccccccc}
\toprule
\multirow{2}{*}{\textbf{Method}}
& \multicolumn{2}{c}{\textbf{MATH500}}
& \multicolumn{2}{c}{\textbf{GSM8K}}
& \multicolumn{2}{c}{\textbf{ARC-C}}
& \multicolumn{2}{c}{\textbf{MMLU}} \\
\cmidrule(l{4pt}r{4pt}){2-3} \cmidrule(l{4pt}r{4pt}){4-5} \cmidrule(l{4pt}r{4pt}){6-7} \cmidrule(l{4pt}r{4pt}){8-9}
&  Latency (ms) & Speedup
&  Latency (ms) & Speedup
&  Latency (ms) & Speedup
&  Latency (ms) & Speedup \\
\midrule
LLaDA2.0-mini &
572.4 & $1.00\times$  &
265.1 & $1.00\times$  &
368.7 & $1.00\times$  &
411.4 & $1.00\times$  \\
\midrule
DES-S  &
394.2 & $1.45\times$  &
161.4 & $1.64\times$  &
300.4 & $1.22\times$  &
312.8 & $1.32\times$  \\
DES-V  &
381.6 & $1.50\times$  &
187.3 & $1.42\times$  &
290.5 & $1.27\times$  &
300.5 & $1.37\times$  \\
\midrule
\rowcolor{gray!30}  ours  &
355.1 & $1.61\times$  &
129.7 & $2.04\times$  &
271.8 & $1.36\times$  &
260.4 & $1.58\times$  \\
\bottomrule
\end{tabular}}
\vspace{-3mm}
\end{table}

\noindent \textbf{\method Brings End-to-End Speedup.}
We further report the end-to-end latency (including MoE latency), together with the corresponding speedup, in Figure~\ref{fig:memory-latency}~(b) to evaluate the practical efficiency of our method. Across four benchmarks, our method achieves \(1.36\times\) to \(2.04\times\) end-to-end latency speedup, with the gain mainly coming from the reduction in MoE latency. In addition, we compare the latency with DES in Table~\ref{tab:latency}, where all methods are compared at comparable accuracy. Our \method achieves the highest speedup on all four benchmarks, whereas DES-S and DES-V only achieve \(1.22\times\) to \(1.64\times\) and \(1.27\times\) to \(1.50\times\) speedup, respectively. Moreover, the overhead of top-p selection and coreset construction is negligible, accounting for only 1.9\% of the total MoE latency on average. These results demonstrate the effectiveness of our method in delivering real inference acceleration.

\noindent \textbf{\method Performs Better at Extreme Expert Compression.}
We further adjust the thresholds of DES-S and DES-V to achieve compression ratios comparable to those of our \method (DES-S$^{*}$ and DES-V$^{*}$ in Table~\ref{tab:llada2-main-1}). We can observe that under similar compression levels, our \method remains nearly lossless, retaining 99.55\% of the original performance with \(p=0.6\) and 97.77\% with \(p=0.5\), whereas DES-S and DES-V suffer from substantial performance degradation. Notably, our \method can compress the number of unique experts to near the practical limit, \textit{i.e.}, the number of experts required per token in the model, which is 8 in LLaDA2.0-mini, while still maintaining strong performance.

\begin{table}[t]
\centering
\caption{Results with larger batch sizes on LLaDA2.0-mini. We report the average latency (Lat., ms), the speedup over the original model (Spd.), the unique expert count across the entire batch (EC), and the memory footprint of uniquely activated MoE parameters (M, GB).}
\label{tab:batch-size}
\vspace{-1.5mm}
\resizebox{0.995\linewidth}{!}{
\setlength{\tabcolsep}{1mm}
\begin{tabular}{lcccccccccccccccc}
\toprule
\multirow{2}{*}{\textbf{Method}}
& \multicolumn{4}{c}{\textbf{MATH500}}
& \multicolumn{4}{c}{\textbf{GSM8K}}
& \multicolumn{4}{c}{\textbf{ARC-C}}
& \multicolumn{4}{c}{\textbf{MMLU}} \\
\cmidrule(l{4pt}r{4pt}){2-5} \cmidrule(l{4pt}r{4pt}){6-9} \cmidrule(l{4pt}r{4pt}){10-13} \cmidrule(l{4pt}r{4pt}){14-17}
&  Lat. & Spd. & EC & M
&  Lat. & Spd. & EC & M
&  Lat. & Spd. & EC & M
&  Lat. & Spd. & EC & M \\
\midrule
\addlinespace
\multicolumn{17}{c}{Batch size \( = 4\)} \\
\midrule
LLaDA2.0-mini &
349.9 & $1.00\times$ & 117.1 & 27  &
167.6 & $1.00\times$ & 117.5 & 27  &
169.7 & $1.00\times$ & 113.4 & 26  &
297.5 & $1.00\times$ & 109.5 & 25  \\
\rowcolor{gray!30}  ours  &
227.3 & $1.54\times$ & 34.4 & 8  &
91.4 & $1.83\times$ & 36.6 & 8  &
98.8 & $1.72\times$ & 38.5 & 9  &
201.9 & $1.47\times$ & 33.1 & 8  \\
\midrule
\addlinespace
\multicolumn{17}{c}{Batch size \( = 8\)} \\
\midrule
LLaDA2.0-mini &
292.9 & $1.00\times$ & 136.4 & 31  &
124.8 & $1.00\times$ & 135.3 & 31  &
152.3 & $1.00\times$ & 118.9 & 27  &
212.3 & $1.00\times$ & 132.9 & 30  \\
\rowcolor{gray!30}  ours  &
188.0 & $1.56\times$ & 44.3 & 10  &
67.7 & $1.84\times$ & 49.1 & 11  &
104.8 & $1.45\times$ & 50.9 & 12  &
149.4 & $1.42\times$ & 44.6 & 10  \\
\bottomrule
\end{tabular}}
\end{table}

\noindent \textbf{\method Remains Effective with Larger Batch Sizes.}
Since larger batch sizes may activate more experts, we further evaluate \method with batch sizes of 4 and 8. As shown in Table~\ref{tab:batch-size}, \method achieves \(1.47\times\) to \(1.83\times\) and \(1.42\times\) to \(1.84\times\) speedup with batch sizes of 4 and 8, respectively, while reducing the unique expert count across the entire batch by 66.05\% to 70.62\% and 57.19\% to 67.52\%. These results further demonstrate the effectiveness and generalizability of \method.

\subsection{Diagnostic Results}
\label{sec:diagnostic-experiments}

We provide comprehensive ablation studies on the cumulative probability threshold \(p\) and block size. Furthermore, we show that our \method achieves a superior performance-efficiency trade-off compared with the baselines.

\begin{table}[t]
\centering
\caption{Ablation on the cumulative probability threshold \(p\) at training and test stages. We evaluate cumulative probability thresholds of $0.4$, $0.5$, $0.6$, $0.7$, and $0.8$ while keeping other settings fixed.} %
\label{tab:ablation-threshold}
\vspace{-1.5mm}
\resizebox{0.995\linewidth}{!}{
\setlength{\tabcolsep}{1mm}
\begin{tabular}{lcccccccc}
\toprule
\multirow{2}{*}{\textbf{Method}} 
& \multicolumn{2}{c}{\textbf{MATH500}} 
& \multicolumn{2}{c}{\textbf{GSM8K}}
& \multicolumn{2}{c}{\textbf{ARC-C}}
& \multicolumn{2}{c}{\textbf{MMLU}} \\ 
\cmidrule(l{4pt}r{4pt}){2-3} \cmidrule(l{4pt}r{4pt}){4-5} \cmidrule(l{4pt}r{4pt}){6-7} \cmidrule(l{4pt}r{4pt}){8-9}
&  Expert Count & Accuracy 
&  Expert Count & Accuracy 
&  Expert Count & Accuracy 
&  Expert Count & Accuracy \\ 
\midrule%
LLaDA2.0-mini & 
70.0 & 72.0\%  & 
71.5 & 90.8\% & 
65.9 & 84.4\%  &
70.6 & 88.6\% \\ 
\midrule
\addlinespace
\multicolumn{9}{c}{\method (\(p^{train} = 0.6\))} \\
\midrule
\rowcolor{gray!30}  ours (\(p^{test} = 0.4\))  &
13.7 & 67.0\% &
13.9 & 86.0\% &
13.1 & 83.7\% &
13.5 & 84.1\% \\
\rowcolor{gray!30}  ours (\(p^{test} = 0.5\))  &
14.0 & 68.8\% &
14.2 & 87.5\% &
13.8 & 85.5\% &
13.7 & 87.0\% \\
\rowcolor{gray!30}  ours (\(p^{test} = 0.6\))  &
16.3 & 71.2\%  &
16.6 & 88.6\%  &
16.9 & 87.1\%  &
16.5 & 87.4\%  \\
\rowcolor{gray!30}  ours (\(p^{test} = 0.7\))  &
19.8 & 73.0\% &
20.1 & 90.2\% &
20.0 & 88.1\% &
20.2 & 87.4\% \\
\rowcolor{gray!30}  ours (\(p^{test} = 0.8\))  &
27.0 & 73.2\% &
28.3 & 90.0\% &
28.0 & 89.2\% &
27.6 & 88.9\% \\
\midrule
\addlinespace
\multicolumn{9}{c}{\method (\(p^{train} = 0.5\))} \\
\midrule
\rowcolor{gray!30}  ours (\(p^{test} = 0.4\))  &
13.0 & 67.2\% &
13.5 & 85.9\% &
12.9 & 83.8\% &
13.1 & 84.1\% \\
\rowcolor{gray!30}  ours (\(p^{test} = 0.5\))  &
13.7 & 69.8\%  &
13.8 & 87.1\%  &
13.9 & 84.7\%  &
13.6 & 86.7\%  \\
\rowcolor{gray!30}  ours (\(p^{test} = 0.6\))  &
16.4 & 70.0\%  &
16.5 & 88.3\%  &
16.1 & 87.0\%  &
16.5 & 87.0\%  \\
\rowcolor{gray!30}  ours (\(p^{test} = 0.7\))  &
19.7 & 73.2\% &
20.2 & 90.2\% &
20.1 & 88.4\% &
19.9 & 87.4\% \\
\rowcolor{gray!30}  ours (\(p^{test} = 0.8\))  &
27.1 & 72.8\% &
28.1 & 90.4\% &
28.0 & 89.0\% &
27.5 & 88.1\% \\
\bottomrule
\end{tabular}}
\end{table}

\noindent \textbf{Ablation on the Cumulative Probability Threshold \(p\).}
We first conduct an ablation study on the cumulative probability threshold \(p\). Specifically, we train two models with \(p^{train} = 0.6\) and \(p^{train} = 0.5\), respectively, and evaluate each tuned model under \(p^{test} \in \{0.4, 0.5, 0.6, 0.7, 0.8\}\). As shown in Table~\ref{tab:ablation-threshold}, when evaluated with the same \(p^{test}\), the two models trained with different \(p^{train}\) achieve similar performance, while still maintaining very low unique expert counts. Moreover, as \(p^{test}\) increases, the unique expert count gradually rises, accompanied by improved performance. Overall, these results demonstrate not only the robustness of our method but also its tunability, allowing the number of activated experts to be flexibly adjusted to different hardware characteristics and application requirements.

\begin{table}[t]
\centering
\caption{Ablation on the block size. The evaluated model is trained with a block size of 32 with \(p^{train} = 0.6\). We set \(p^{test} = 0.6\) here. We evaluate block sizes of $8$, $16$, $24$, and $32$ while keeping all other hyperparameters fixed.} %
\label{tab:ablation-blocksize}
\vspace{-1.5mm}
\resizebox{0.995\linewidth}{!}{
\setlength{\tabcolsep}{1mm}
\begin{tabular}{lcccccccc}
\toprule
\multirow{2}{*}{\textbf{Method}} 
& \multicolumn{2}{c}{\textbf{MATH500}} 
& \multicolumn{2}{c}{\textbf{GSM8K}}
& \multicolumn{2}{c}{\textbf{ARC-C}}
& \multicolumn{2}{c}{\textbf{MMLU}} \\ 
\cmidrule(l{4pt}r{4pt}){2-3} \cmidrule(l{4pt}r{4pt}){4-5} \cmidrule(l{4pt}r{4pt}){6-7} \cmidrule(l{4pt}r{4pt}){8-9}
&  Expert Count & Accuracy 
&  Expert Count & Accuracy 
&  Expert Count & Accuracy 
&  Expert Count & Accuracy \\ 
\midrule
\addlinespace
\multicolumn{9}{c}{Block size \( = 32\)} \\
\midrule
LLaDA2.0-mini & 
70.0 & 72.0\%  & 
71.5 & 90.8\% & 
65.9 & 84.4\%  &
70.6 & 88.6\% \\ 
\rowcolor{gray!30}  ours  &
16.3 & 71.2\%  &
16.6 & 88.6\%  &
16.9 & 87.1\%  &
16.5 & 87.4\%  \\
\midrule
\addlinespace
\multicolumn{9}{c}{Block size \( = 24\)} \\
\midrule
LLaDA2.0-mini & 
60.7 & 73.6\%  &
62.2 & 91.0\%  &
57.8 & 82.3\%  &
61.6 & 86.9\% \\
\rowcolor{gray!30}  ours  &
16.2 & 68.6\%  &
17.4 & 87.4\%  &
16.1 & 87.9\%  &
15.8 & 85.3\% \\
 \midrule
\addlinespace
\multicolumn{9}{c}{Block size \( = 16\)} \\
\midrule
LLaDA2.0-mini & 
48.6 & 70.2\%  &
49.8 & 90.5\%  &
47.0 & 86.9\%  &
49.6 & 88.5\% \\
\rowcolor{gray!30}  ours  &
14.8 & 70.0\%  &
16.7 & 89.9\%  &
15.1 & 88.1\%  &
14.9 & 87.4\% \\
\midrule
\addlinespace
\multicolumn{9}{c}{Block size \( = 8\)} \\
\midrule
LLaDA2.0-mini & 
31.5 & 71.0\%  &
32.2 & 90.1\%  &
31.1 & 87.7\%  &
32.3 & 87.4\% \\
\rowcolor{gray!30}  ours  &
14.1 & 70.2\%  &
16.1 & 88.1\%  &
14.2 & 87.2\%  &
14.1 & 86.7\% \\
\bottomrule
\end{tabular}}
\end{table}

\noindent \textbf{Ablation on the Block Size.}
We further conduct an ablation study on the block size, with the results summarized in Table~\ref{tab:ablation-blocksize}. Specifically, we evaluate the model under block sizes of 8, 16, 24, and 32. The results show that our \method consistently compresses the unique expert count while maintaining strong performance across different block sizes. Specifically, under block sizes of 32, 24, 16, and 8, our \method reduces the average unique expert count by 76.15\%, 72.97\%, 68.46\%, and 53.97\%, while retaining 99.55\%, 98.62\%, 99.79\%, and 98.81\% of the original performance, respectively. The consistent efficiency improvement without a performance drop across different block sizes further demonstrates the effectiveness of our method.

\begin{figure*}[t]
\centering
\includegraphics[width=\linewidth]{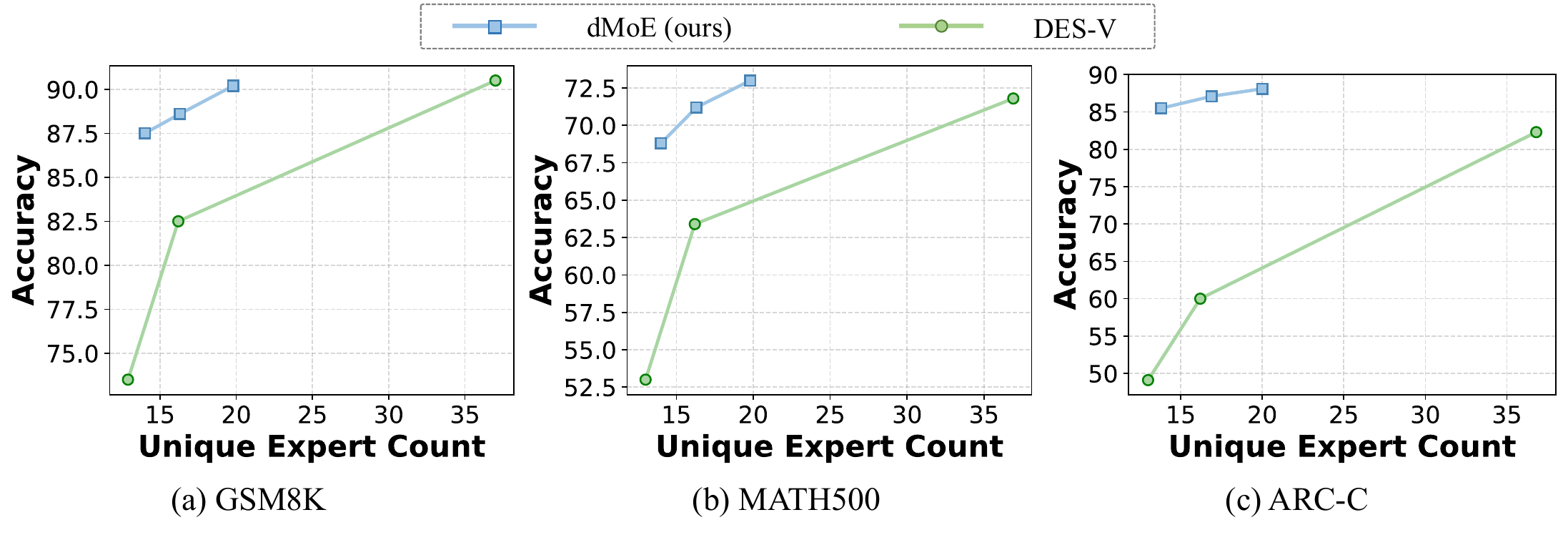}  \\
\vspace{-5mm}
\caption{Comparison of the performance-efficiency trade-off between our method and baselines. We report the results of GSM8K (a), MATH500 (b), and ARC-C (c).}
\label{fig:tradeoff}
\vspace{-3mm}
\end{figure*}

\noindent \textbf{\method Reaches Superior Performance-Efficiency Trade-off.}
We present the performance--efficiency trade-off between our \method and the baselines. The results of our \method are taken from Table~\ref{tab:ablation-threshold}. For DES, we compare against its stronger variant, DES-V, and obtain its trade-off points by varying the threshold. As shown in Figure~\ref{fig:tradeoff}, our \method achieves better performance with fewer unique experts, leading to a superior performance--efficiency trade-off.

\section{Conclusion}
\label{sec:conclusion}

In this work, we introduce \method, a novel strategy for block-level expert routing in MoE dLLMs. \method enables aggressive expert compression while incurring almost no performance degradation. Comprehensive evaluations across diverse benchmarks demonstrate the effectiveness of our method. More broadly, this work lays a foundation for block-level routing in MoE dLLMs and opens up a promising direction for improving their efficiency.

\section*{Acknowledgements}
This project is supported by the Ministry of Education, Singapore, under the Academic Research Fund Tier~1 (FY2026, WBS: A-8004345-00-00),  the National Research Foundation, Singapore, and Cyber Security Agency of Singapore under its National Cybersecurity R\&D Programme and CyberSG R\&D Cyber Research Programme Office (Award: CRPO-GC1-NTU-002), as well as NUS IT's High Performance Cluster.

{
\small
\bibliographystyle{unsrt}
{\small\bibliography{main.bib}}
}

\clearpage
\appendix
\onecolumn
\setcounter{figure}{0}
\setcounter{table}{0}
\renewcommand{\thefigure}{A\arabic{figure}}
\renewcommand{\thetable}{A\arabic{table}}
\section*{Appendix}
\label{apx:apx}

In Appendix~\ref{apx:add-prelim}, we provide additional preliminaries. In Appendix~\ref{apx:limit-future}, we discuss limitations and future work. We further provide the impact statement in Appendix~\ref{apx:impact} and license statement in Appendix~\ref{apx:license-content-info}, and computing resources in Appendix~\ref{apx:resource}.

\vspace{-0.2cm}
\startcontents[appendices]
\printcontents[appendices]{l}{1}{\setcounter{tocdepth}{3}}

\section{Additional Preliminaries}
\label{apx:add-prelim}

\subsection{Unique Expert Count Across Denoising Steps}
\label{apx:expert-count}

For Figure~\ref{fig:observations}~(b), following the setting in Observation B, we record the unique expert count at each denoising step and plot its frequency distribution as a histogram. The unique expert count ranges from approximately 45 to 95, indicating nearly twofold variation and substantial fluctuation across denoising steps. We further present a representative sequence of consecutive denoising steps in Figure~\ref{fig:expert-count-steps}, where the unique expert count exhibits large periodic fluctuations, further supporting Observation B.

\begin{figure}[ht]
\centering
\begin{tikzpicture}
\begin{axis}[
    width=\linewidth,
    height=4.5cm,
    xmin=-1, xmax=40,
    ymin=40, ymax=90,
    xtick={0,5,...,35},
    ytick={40,50,...,90},
    xlabel={Denoising step},
    ylabel={Unique expert count},
    label style={font=\small},
    tick label style={font=\footnotesize},
    grid=major,
    grid style={dashed, gray!30},
]
\addplot[color=map_blue, line width=1pt, mark=square*, mark size=1.5pt] coordinates {
    (0,49) (1,55) (2,66) (3,64) (4,66) (5,67) (6,65) (7,69) (8,70) (9,74)
    (10,75) (11,79) (12,80) (13,82) (14,82) (15,85) (16,80)
    (17,53) (18,59) (19,64) (20,61) (21,65) (22,67) (23,67) (24,70) (25,72)
    (26,74) (27,71) (28,72) (29,76) (30,76) (31,76)
    (32,53) (33,57) (34,56) (35,64) (36,58) (37,56) (38,59) (39,62)
};
\end{axis}
\end{tikzpicture}
\vspace{-4mm}
\caption{Unique expert count at each step over a representative sequence of consecutive denoising steps.}
\label{fig:expert-count-steps}
\end{figure}

\section{Limitations \& Future Work}
\label{apx:limit-future}

Our method is, in principle, applicable to any MoE dLLMs. In this work, we focus our evaluation on the language modality, but the same idea can be naturally extended to other modalities, such as image and video, and to downstream tasks including visual question answering~\citep{feng2026reasonmap,shao2025tokens,zhu2025obs} and visual reasoning~\citep{feng2026rewardmap,tao2025omnizip,jin2025mergemix,ai2026pasa,du2025whichheads,wang2026earlytom}. More broadly, our results suggest that block-level expert routing is a promising direction for future MoE dLLMs. Looking ahead, this direction can be pushed further toward more extreme compression, for example, by encouraging all tokens within a block to share the same expert group or by jointly reducing computation cost through selecting fewer experts for each token.

\section{Impact Statement}
\label{apx:impact}

This work studies efficient inference for MoE dLLMs by reducing the number of uniquely activated experts during decoding. By implementing block-level expert routing, the proposed method reduces memory overhead, lowers inference cost, and alleviates the memory bottleneck in MoE dLLMs. These improvements make large-scale model deployment more practical, especially in resource-constrained, memory-sensitive, or latency-sensitive settings. More broadly, this work contributes to improving the efficiency and accessibility of dLLMs in real-world applications. Furthermore, the method does not introduce additional ethical or societal concerns.
\section{License Statement}
\label{apx:license-content-info}

The model and datasets used in this paper are publicly available, and all experiments are conducted in compliance with their respective licenses. The specific licenses for the model and datasets are listed below.
\begin{itemize}
    \item LLaDA2.0-mini\footnote{\url{https://huggingface.co/inclusionAI/LLaDA2.0-mini}.} \dotfill Apache 2.0 License
    \item GSM8K\footnote{\url{https://huggingface.co/datasets/openai/gsm8k}.} \dotfill MIT License
    \item MATH500\footnote{\url{https://huggingface.co/datasets/HuggingFaceH4/MATH-500}} \dotfill MIT License
    \item ARC-C\footnote{\url{https://huggingface.co/datasets/allenai/ai2_arc}} \dotfill cc-by-sa-4.0 License
    \item MMLU\footnote{\url{https://huggingface.co/datasets/cais/mmlu}} \dotfill MIT License
    \item TruthfulQA\footnote{\url{https://huggingface.co/datasets/truthfulqa/truthful_qa}} \dotfill Apache 2.0 License
    \item QASPER\footnote{\url{https://huggingface.co/datasets/allenai/qasper}} \dotfill cc-by-4.0 License
\end{itemize}
To ensure fair and reproducible evaluation, we follow the official documentation and recommended practices of each model when implementing the inference procedures.
\section{Computing Resources}
\label{apx:resource}

The experiments were performed on a server equipped with two AMD EPYC 9654 96-Core processors, 1.5 TiB of system memory, and four NVIDIA H100 GPUs with 80 GB VRAM each. We employed data parallelism to speed up evaluation. Each evaluation run was completed within 12 hours, while each training run finished within 144 hours.

\newpage
\section*{NeurIPS Paper Checklist}

\begin{enumerate}

\item {\bf Claims}
    \item[] Question: Do the main claims made in the abstract and introduction accurately reflect the paper's contributions and scope?
    \item[] Answer: \answerYes{} %
    \item[] Justification: The abstract and introduction accurately state our contributions and scope, and these claims are supported by the experimental results in Section~\ref{sec:experiments}.

\item {\bf Limitations}
    \item[] Question: Does the paper discuss the limitations of the work performed by the authors?
    \item[] Answer: \answerYes{} %
    \item[] Justification: We discuss the limitations of our work and future directions in Appendix~\ref{apx:limit-future}.

\item {\bf Theory assumptions and proofs}
    \item[] Question: For each theoretical result, does the paper provide the full set of assumptions and a complete (and correct) proof?
    \item[] Answer: \answerNA{} %
    \item[] Justification: The paper does not include theoretical results.

    \item {\bf Experimental result reproducibility}
    \item[] Question: Does the paper fully disclose all the information needed to reproduce the main experimental results of the paper to the extent that it affects the main claims and/or conclusions of the paper (regardless of whether the code and data are provided or not)?
    \item[] Answer: \answerYes{} %
    \item[] Justification: All information needed to reproduce the main results, including training data construction, training details, baselines, and inference and evaluation settings, is provided in Section~\ref{sec:experimental-setup}. Our code is also publicly available at \url{https://github.com/fscdc/dMoE}.

\item {\bf Open access to data and code}
    \item[] Question: Does the paper provide open access to the data and code, with sufficient instructions to faithfully reproduce the main experimental results, as described in supplemental material?
    \item[] Answer: \answerYes{} %
    \item[] Justification: Our code is publicly available at \url{https://github.com/fscdc/dMoE}, with instructions for environment setup, training, and evaluation. The training data are constructed from public datasets via self-distillation, as described in Section~\ref{sec:experimental-setup}.

\item {\bf Experimental setting/details}
    \item[] Question: Does the paper specify all the training and test details (e.g., data splits, hyperparameters, how they were chosen, type of optimizer) necessary to understand the results?
    \item[] Answer: \answerYes{} %
    \item[] Justification: The training and test details, including data construction, hyperparameters, and inference and evaluation settings, are provided in Section~\ref{sec:experimental-setup}.

\item {\bf Experiment statistical significance}
    \item[] Question: Does the paper report error bars suitably and correctly defined or other appropriate information about the statistical significance of the experiments?
    \item[] Answer: \answerYes{} %
    \item[] Justification: Sufficient information about experiments has been discussed in Section~\ref{sec:experiments}.

\item {\bf Experiments compute resources}
    \item[] Question: For each experiment, does the paper provide sufficient information on the computer resources (type of compute workers, memory, time of execution) needed to reproduce the experiments?
    \item[] Answer: \answerYes{} %
    \item[] Justification: The hardware, memory, and running time of our experiments are reported in Appendix~\ref{apx:resource}.
    
\item {\bf Code of ethics}
    \item[] Question: Does the research conducted in the paper conform, in every respect, with the NeurIPS Code of Ethics \url{https://neurips.cc/public/EthicsGuidelines}?
    \item[] Answer: \answerYes{} %
    \item[] Justification: This research follows the NeurIPS Code of Ethics properly.

\item {\bf Broader impacts}
    \item[] Question: Does the paper discuss both potential positive societal impacts and negative societal impacts of the work performed?
    \item[] Answer: \answerYes{} %
    \item[] Justification: We discuss the societal impacts of our work in Appendix~\ref{apx:impact}.
    
\item {\bf Safeguards}
    \item[] Question: Does the paper describe safeguards that have been put in place for responsible release of data or models that have a high risk for misuse (e.g., pre-trained language models, image generators, or scraped datasets)?
    \item[] Answer: \answerNA{} %
    \item[] Justification: The paper poses no such risks.

\item {\bf Licenses for existing assets}
    \item[] Question: Are the creators or original owners of assets (e.g., code, data, models), used in the paper, properly credited and are the license and terms of use explicitly mentioned and properly respected?
    \item[] Answer: \answerYes{} %
    \item[] Justification: All models and datasets used in this paper are properly cited, and their licenses are listed in Appendix~\ref{apx:license-content-info}.

\item {\bf New assets}
    \item[] Question: Are new assets introduced in the paper well documented and is the documentation provided alongside the assets?
    \item[] Answer: \answerYes{} %
    \item[] Justification: We release our code and model at \url{https://github.com/fscdc/dMoE}, together with documentation on environment setup, usage, training, and evaluation.

\item {\bf Crowdsourcing and research with human subjects}
    \item[] Question: For crowdsourcing experiments and research with human subjects, does the paper include the full text of instructions given to participants and screenshots, if applicable, as well as details about compensation (if any)? 
    \item[] Answer: \answerNA{} %
    \item[] Justification: This work does not involve crowdsourcing nor research with human subjects.

\item {\bf Institutional review board (IRB) approvals or equivalent for research with human subjects}
    \item[] Question: Does the paper describe potential risks incurred by study participants, whether such risks were disclosed to the subjects, and whether Institutional Review Board (IRB) approvals (or an equivalent approval/review based on the requirements of your country or institution) were obtained?
    \item[] Answer: \answerNA{} %
    \item[] Justification: This work does not involve crowdsourcing nor research with human subjects.

\item {\bf Declaration of LLM usage}
    \item[] Question: Does the paper describe the usage of LLMs if it is an important, original, or non-standard component of the core methods in this research? Note that if the LLM is used only for writing, editing, or formatting purposes and does \emph{not} impact the core methodology, scientific rigor, or originality of the research, declaration is not required.
    \item[] Answer: \answerNA{} %
    \item[] Justification: The core method development in this research does not involve LLMs as any important, original, or non-standard components.

\end{enumerate}

\end{document}